\documentclass[10pt,leqno]{amsart}
\usepackage{graphicx}
\usepackage{indentfirst,csquotes}
\usepackage{standalone}
\usepackage{graphicx}
\usepackage{indentfirst,csquotes}
\usepackage{amssymb,amsthm,amsmath}
\usepackage{xcolor,paralist,hyperref,fancyhdr,etoolbox} 
\usepackage{braket}
\usepackage{subcaption} 
\usepackage{booktabs} 
\usepackage{tikz}
\usepackage{pgfplots}
\usepackage{lmodern}
\usepackage{import}
\usepackage{algorithm}
\usepackage{algpseudocode}
\usepackage{booktabs}
\usepackage{mdframed}
\usepackage{tcolorbox}
\usepackage[table]{xcolor}
\usetikzlibrary{automata, positioning}
\usetikzlibrary{shapes.geometric, arrows, positioning}

\everymath=\expandafter{\the\everymath\displaystyle}

\tikzstyle{block} = [rectangle, draw, text centered, rounded corners, minimum height=2em]
\tikzstyle{line} = [draw, -stealth, thick]
\tikzstyle{cloud} = [ellipse, draw, text centered, minimum height=2em, thick]
\tikzstyle{dashedcloud} = [ellipse, draw, dashed, text centered, minimum height=2em, thick]
\usetikzlibrary{positioning, arrows.meta, shapes.geometric, decorations.pathreplacing}

\tikzstyle{startstop} = [rectangle, rounded corners, minimum width=1.5cm, minimum height=0.5cm,text centered, draw=black, fill=red!30]
\tikzstyle{io} = [trapezium, trapezium left angle=70, trapezium right angle=110, minimum width=1cm, minimum height=0.5cm, text centered, draw=black, fill=blue!30]

\tikzstyle{process} = [rectangle, minimum width=3cm, minimum height=0.5cm, text centered, draw=black, fill=orange!30]
\tikzstyle{decision} = [diamond, minimum width=0.5cm, minimum height=0.1cm, text centered, draw=black, fill=green!30]
\tikzstyle{process2} = [rectangle, minimum width=1cm, minimum height=0.5cm, text centered, draw=black, fill=orange!30]
\tikzstyle{arrow} = [thick,->,>=stealth]
\usetikzlibrary{shapes,shadows,arrows,positioning,angles,quotes}
\tikzset{My Arrow Style/.style={single arrow, fill=black!15, anchor=base, align=center,text width=2.3cm}}
\tikzstyle{arrow} = [thick,->,>=stealth]
\usetikzlibrary{calc,trees,positioning,arrows,fit,shapes,calc}

\newtheorem{theorem}{Theorem}
\newtheorem{definition}[theorem]{Definition}

\hypersetup{
  colorlinks=true,
  linkcolor=blue,    
  citecolor=blue,    
  filecolor=blue,    
  urlcolor=blue      
}

\begin{document}

\title{Translating the Rashomon Effect to Sequential Decision-Making Tasks}
\author[]{Dennis Gross, J\o rn Eirik Betten, Helge Spieker}
\maketitle


\begin{abstract}
The \emph{Rashomon effect} describes the phenomenon where multiple models trained on the same data produce identical predictions while differing in which features they rely on internally.
This effect has been studied extensively in classification tasks, but not in sequential decision-making, where an agent learns a policy to achieve an objective by taking actions in an environment.
In this paper, we translate the Rashomon effect to sequential decision-making. We define it as multiple policies that exhibit identical behavior, visiting the same states and selecting the same actions, while differing in their internal structure, such as feature attributions.
Verifying identical behavior in sequential decision-making differs from classification.
In classification, predictions can be directly compared to ground-truth labels.
In sequential decision-making with stochastic transitions, the same policy may succeed or fail on any single trajectory due to randomness.
We address this using formal verification methods that construct and compare the complete probabilistic behavior of each policy in the environment.
Our experiments demonstrate that the Rashomon effect exists in sequential decision-making.
We further show that ensembles constructed from the \emph{Rashomon set} exhibit greater robustness to distribution shifts than individual policies.
Additionally, permissive policies derived from the Rashomon set reduce computational requirements for verification while maintaining optimal performance.
\end{abstract}


\section{Introduction}
\emph{Supervised learning} is a type of machine learning where a model learns from labeled examples~\cite{DBLP:conf/icitjo/AbdiMWKASASA25,DBLP:conf/glvlsi/Hu22}.
A \emph{classification task} is the problem of predicting which class an input belongs to, and supervised learning can train models to perform this task~\cite{goodfellow2016deep}.

\emph{The Rashomon effect in classification tasks refers to the phenomenon where multiple models trained on the same dataset produce identical predictions while differing in which features they emphasize or how they process inputs internally}~\cite{breiman2001statistical,fisher_all_2019,semenova_existence_2022,Xin2022,muller_empirical_2023,Laberge2023,Andersen2023}.
These internal differences can be revealed through \emph{explainable machine learning methods} such as \emph{saliency maps}, which assign an importance score to each feature~\cite{DBLP:conf/aiccsa/DakhliB24}.

The trained models form the \emph{Rashomon set}~\cite{Xin2022,Laberge2023}, which can be combined into an \emph{ensemble}~\cite{DBLP:conf/atva/Gross0PR20} whose aggregated predictions yield a single classification output.
Such ensembles often exhibit greater robustness to distribution shifts than randomly selecting a single model from this set~\cite{Xin2022,Laberge2023,DBLP:conf/atva/Gross0PR20}.

\emph{However}, no prior work has extended the Rashomon effect to \emph{sequential decision-making tasks}~\cite{DBLP:journals/jair/NashedMGZ25}, in which an agent learns a policy to achieve an objective by taking actions and receiving feedback in the form of rewards and state observations from the environment~\cite{sutton2018reinforcement}. A policy is \emph{memoryless} if it depends only on the current state (see Figure~\ref{fig:rl}).

Since action selection at each state is essentially a classification problem, policies can be trained on a labeled dataset via behavioral cloning, a supervised learning approach in which a policy learns to imitate expert demonstrations~\cite{DBLP:conf/icoin/ShinK21,DBLP:conf/icaart/SchmidlS024,DBLP:journals/corr/abs-2204-05618}.

\emph{In sequential decision-making}, a policy's observable behavior concerning an objective is fully characterized by its \emph{induced discrete-time Markov chain (induced DTMC)}: the reachable states it can visit and the resulting transitions between states.
Two policies that produce identical induced DTMCs for a specified objective are observationally indistinguishable; they navigate the environment in exactly the same way.
\emph{Probabilistic model checking}~\cite{baier2008principles} enables the exact construction and comparison of these induced DTMCs, in which objectives can be expressed as PCTL temporal logic properties, such as ``reach the goal with probability 1.''
The tool \emph{COOL-MC}~\cite{DBLP:conf/setta/Gross22} applies this technique in practice: given a trained policy, a formal environment model, and a property specification, it constructs the induced Markov chain and verifies whether the property is satisfied.
The induced Markov chain captures all policy-environment interactions relevant to this property.

\emph{Prior work} in probabilistic model checking has studied bisimulation, which formalizes when two systems exhibit identical observable behavior~\cite{DBLP:journals/iandc/LarsenS91,ferns2011bisimulation,delgrange2023wae}.
While bisimulation can identify policies with equivalent behavior, it does not address whether these policies were trained on the same data or whether they differ in their internal structure.
In contrast, the Rashomon effect specifically concerns policies trained on identical data that behave identically yet differ internally.
By situating behaviorally equivalent policies within the Rashomon framework, we can identify which policies form a Rashomon set and which do not, potentially providing deeper insights for explainable machine learning in sequential decision-making.

\emph{In this paper}, we extend the Rashomon effect to sequential decision-making and show that the Rashomon set of policies can exhibit greater robustness to shifts in the environment distribution than randomly selecting a single one from this set.

\begin{figure}[t]
    \centering
    \scalebox{0.95}{
    \tikzset{
    input node/.style={
        rectangle,
        rounded corners=3pt,
        draw=black,
        line width=0.8pt,
        fill=gray!15,
        minimum width=1.6cm,
        minimum height=0.8cm,
        align=center,
        font=\sffamily\small
    },
    input container/.style={
        rectangle,
        rounded corners=3pt,
        draw=black,
        line width=0.8pt,
        fill=gray!15,
        minimum width=2.4cm,
        minimum height=3.2cm,
        align=center
    },
    inner box/.style={
        rectangle,
        rounded corners=2pt,
        draw=black,
        line width=0.5pt,
        fill=white,
        minimum width=2cm,
        minimum height=0.65cm,
        align=center,
        font=\sffamily\footnotesize
    },
    classifier card/.style={
        rectangle,
        rounded corners=3pt,
        draw=black,
        line width=0.8pt,
        fill=white,
        minimum width=3.2cm,
        minimum height=2.2cm,
        align=center
    },
    classifier header/.style={
        rectangle,
        rounded corners=2pt,
        draw=black,
        line width=0.5pt,
        fill=gray!15,
        minimum width=2.8cm,
        minimum height=0.55cm,
        align=center,
        font=\sffamily\small
    },
    feature tag/.style={
        rectangle,
        rounded corners=2pt,
        draw=black,
        line width=0.5pt,
        fill=white,
        minimum width=2.6cm,
        minimum height=0.55cm,
        align=center,
        font=\sffamily\scriptsize
    },
    output label/.style={
        rectangle,
        rounded corners=3pt,
        draw=black,
        line width=0.8pt,
        fill=gray!15,
        minimum width=2cm,
        minimum height=0.8cm,
        align=center,
        font=\sffamily\small
    },
    arrow/.style={
        -{Stealth[length=2mm, width=1.5mm]},
        line width=0.6pt,
        color=black
    },
    connector/.style={
        circle,
        fill=black,
        inner sep=1pt
    }
}

\begin{tikzpicture}


\node[input node] (input1) at (0,0) {Input};

\node[connector] (conn1) at (1, 0) {};

\node[classifier card] (c1l) at (3.4, 2.2) {};
\node[classifier header] at (3.4, 2.8) {Classifier 1};
\node[feature tag] at (3.4, 2) {Features: $x > y > z$};

\node[classifier card] (c2l) at (3.4, 0) {};
\node[classifier header] at (3.4, 0.6) {Classifier 2};
\node[feature tag] at (3.4, -0.2) {Features: $z > x > y$};

\node[classifier card] (c3l) at (3.4, -2.2) {};
\node[classifier header] at (3.4, -1.6) {Classifier 3};
\node[feature tag] at (3.4, -2.4) {Features: $y > z > x$};

\node[output label] (l1) at (6.2, 2.2) {Prediction A};
\node[output label] (l2) at (6.2, 0) {Prediction A};
\node[output label] (l3) at (6.2, -2.2) {Prediction A};

\draw[arrow] (input1.east) -- (conn1);
\draw[arrow] (conn1) -- (c1l.west);
\draw[arrow] (conn1) -- (c2l.west);
\draw[arrow] (conn1) -- (c3l.west);

\draw[arrow] (c1l.east) -- (l1.west);
\draw[arrow] (c2l.east) -- (l2.west);
\draw[arrow] (c3l.east) -- (l3.west);


\node[input container] (inputblock) at (9.8, 0) {};
\node[font=\sffamily\small] at (9.8, 1.1) {Input};
\node[inner box] (mdp) at (9.8, 0.3) {Environment};
\node[inner box] (prop) at (9.8, -0.55) {Objective};

\node[connector] (conn2) at (11.3, 0) {};

\node[classifier card] (c1r) at (13.7, 2.2) {};
\node[classifier header] at (13.7, 2.8) {COOL-MC (Policy 1)};
\node[feature tag] at (13.7, 2) {Features: $x > y > z$};

\node[classifier card] (c2r) at (13.7, 0) {};
\node[classifier header] at (13.7, 0.6) {COOL-MC (Policy 2)};
\node[feature tag] at (13.7, -0.2) {Features: $z > x > y$};

\node[classifier card] (c3r) at (13.7, -2.2) {};
\node[classifier header] at (13.7, -1.6) {COOL-MC (Policy 3)};
\node[feature tag] at (13.7, -2.4) {Features: $y > z > x$};

\node[output label] (s1) at (16.5, 2.2) {MC A};
\node[output label] (s2) at (16.5, 0) {MC A};
\node[output label] (s3) at (16.5, -2.2) {MC A};

\draw[arrow] (inputblock.east) -- (conn2);
\draw[arrow] (conn2) -- (c1r.west);
\draw[arrow] (conn2) -- (c2r.west);
\draw[arrow] (conn2) -- (c3r.west);

\draw[arrow] (c1r.east) -- (s1.west);
\draw[arrow] (c2r.east) -- (s2.west);
\draw[arrow] (c3r.east) -- (s3.west);

\end{tikzpicture}
    }
    \caption{An example comparison of the Rashomon effect in classification (left) and sequential decision-making (right). In both cases, classification models and COOL-MC with policies produce identical outputs (predictions/induced DTMCs) while differing internally (feature attributions).}
    \label{fig:classifiers}
\end{figure}
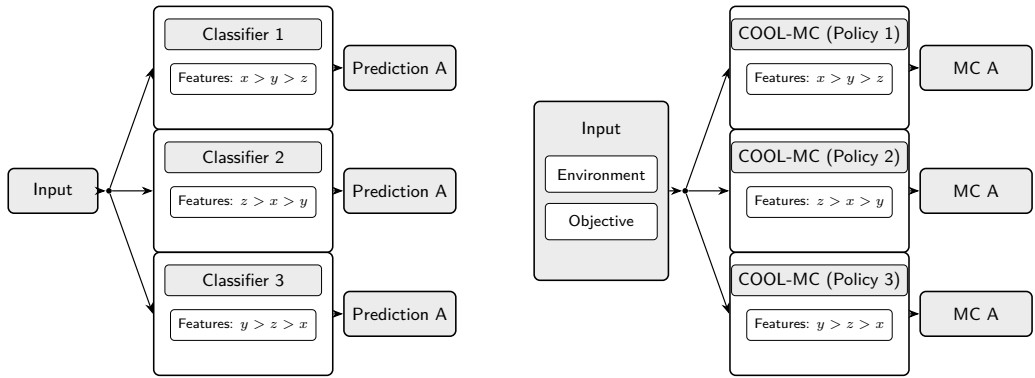

In more detail, \emph{the Rashomon effect in sequential decision-making} describes the observation that multiple policies trained on the same data select the same action in every state of an environment concerning a property, thereby producing identical induced DTMCs, while differing in their internal structure as defined by a user-specified metric, such as feature attributions.
This is analogous to the classification Rashomon effect, where models make identical predictions while differing internally; here, identical induced DTMCs concerning a specified objective are synonymous with identical predictions (see visual comparison in Figure~\ref{fig:classifiers}).

The set of policies exhibiting this effect forms the \emph{Rashomon set}.
In contrast, policies that are identical with respect to both observable behavior and internal structure under the specified metric do not form a Rashomon set.

In classification and regression, with regression viewed as a special case of classification, verifying identical behavior is straightforward. One can directly compare model predictions with the ground-truth labels.
In sequential decision-making with stochastic transitions, this simple comparison to the ground truth is not straightforward, as the same policy may succeed or fail on any single trajectory due to the stochastic nature of the environment dynamics.
Probabilistic model checking resolves this by explicitly constructing and comparing the induced DTMCs, whereas user-specified explainable machine learning methods verify that the internal structures differ.

\emph{Our experiments} show that the Rashomon effect exists in sequential decision-making.
Importantly, we show that not all policies with identical induced DTMCs constitute a Rashomon set: only those that additionally differ in their internal structure under the specified metric.
Furthermore, we show that these internal differences have practical consequences.
Under distribution shifts, such as changes to the environment not seen during training, combining policies from the Rashomon set may be more robust in achieving the objective than selecting any individual policy.

Our \textbf{main contributions} are the translation and verification of the Rashomon effect in sequential decision-making, and the demonstration that ensembles built out of the Rashomon set can outperform individual policies under distribution shifts.

\section{Related Work}
In his 2001 paper, Leo Breiman defines the Rashomon effect as the observation that many different models can achieve similar performance, each offering a different perspective on which variables are important, despite fitting the data equally well.
The key insight is that these models make the same predictions while differing internally.
We translate this concept to sequential decision-making tasks, where ``same predictions'' becomes ``same induced Markov chain for a specified property.''

There are two main contexts in which the Rashomon effect has been studied.
One concerns the existence and selection of simpler, more interpretable models in the Rashomon set for a given dataset \cite{fisher_all_2019,semenova_existence_2022,Xin2022}, the other the impacts on explainability of models in the Rashomon set \cite{muller_empirical_2023,Laberge2023,Andersen2023}.

Spieker et al. provide the first empirical quantification of the Rashomon effect for action prediction in autonomous driving. They train Rashomon sets of gradient boosting models and graph neural networks, finding that models with similar validation performance produce significantly divergent explanations for the same predictions~\cite{DBLP:journals/corr/abs-2509-03169}. 

In sequential decision-making, it is well known that for a specified objective, there can exist multiple optimal policies to achieve it.
However, these multiple optimal policies typically represent genuinely different strategies that visit different states or take different actions.
Our work identifies a distinct phenomenon: policies that are not merely optimal for the same objective, but that produce \emph{identical observable behavior} while differing in their internal representations.
By using behavioral cloning to train multiple policies on an identical expert dataset and verifying that they induce identical DTMCs, we isolate the Rashomon effect, in which all differences are internal.

For the policy property verification and induced Markov chain construction, we use COOL-MC~\cite{DBLP:conf/setta/Gross22}, and for the expert dataset creation, the Storm model checker~\cite{DBLP:journals/sttt/HenselJKQV22}.

Bisimulation formalizes when two systems exhibit identical observable behavior~\cite{DBLP:journals/iandc/LarsenS91,ferns2011bisimulation,delgrange2023wae}. Our work is related but asks a different question. Bisimulation determines whether two systems behave identically. We start from systems (policies) that we have verified to behave identically via their induced DTMCs, and then investigate whether they differ internally in their feature~attributions.

\section{Background}
First, we outline how sequential decision-making tasks can be formally represented as \emph{Markov decision processes (MDPs)}, and how memoryless stochastic policies and their properties can be verified with this method.
Second, we introduce behavioral cloning for training sequential decision-making policies.

\subsection{Probabilistic Systems}
A \textit{probability distribution} over a set $X$ is a function $\mu \colon X \rightarrow [0,1]$ with $\sum_{x \in X} \mu(x) = 1$. The set of all distributions on $X$ is $Distr(X)$.

\begin{definition}[MDP]\label{def:mdp}
A \emph{MDP} is a tuple $M = (S,s_0,Act,Tr, rew, AP,L)$
where $S$ is a finite, nonempty set of states; $s_0 \in S$ is an initial state; $Act$ is a finite set of actions; $Tr\colon S \times Act \rightarrow Distr(S)$ is a partial probability transition function and $Tr(s,a,s')$ denotes the probability of transitioning from state $s$ to state $s'$ when action $a$ is taken;
$rew \colon S \times Act \rightarrow \mathbb{R}$ is a reward~function;
$AP$ is a set of atomic propositions;
$L \colon  S \rightarrow 2^{AP}$ is a labeling~function.
\end{definition}

We represent each state $s \in S$ as a vector of $d$ integer features $(f_1, \dots, f_d)$, where $f_i \in \mathbb{Z}$.
The available actions in $s \in S$ are $Act(s) = \{a \in Act \mid Tr(s,a) \neq \bot\}$ where $Tr(s, a) \neq \bot$ is defined as action $a$ at state $s$ does not have a transition (action $a$ is not available in state $s$).
In our setting, we assume that all actions are available at all states.

\begin{definition}[DTMC]\label{def:dtmc}
A \emph{discrete-time Markov chain (DTMC)} is a tuple $D = (S, s_0, Tr, AP, L)$
where $S$, $s_0$, $AP$, and $L$ are as in Definition~\ref{def:mdp}, 
and $Tr \colon S \rightarrow Distr(S)$ is a probability transition~function.
\end{definition}

\begin{definition}
    A \emph{memoryless deterministic policy $\pi$} for an MDP $M$ is a function $\pi \colon S \rightarrow Act$ that maps a state $s \in S$ to action $a \in Act$.
\end{definition}
Applying a policy $\pi$ to an MDP $M$ yields an \emph{induced DTMC (Discrete-Time Markov Chain)} $D$ where all non-determinism is resolved.
The induced DTMC fully characterizes the observable behavior of the policy: the states visited, the transitions taken, and the probabilities of all outcomes.
Two policies that produce identical induced DTMCs are observationally indistinguishable.

\begin{figure}[]
\centering
\scalebox{0.25}{
    \tikzset{
    process box/.style={
        rectangle,
        rounded corners=3pt,
        draw=black,
        line width=0.8pt,
        fill=white,
        minimum width=2.5cm,
        minimum height=1cm,
        align=center,
        font=\sffamily\small
    },
    arrow/.style={
        -{Stealth[length=2mm, width=1.5mm]},
        line width=0.6pt,
        color=black
    },
    label style/.style={
        font=\sffamily\footnotesize
    }
}

\begin{tikzpicture}

\node[process box] (policy) {Policy};
\node[process box, below=1.8cm of policy] (env) {Environment};

\draw[arrow] ([xshift=-0.4cm]policy.south) -- node[label style, left] {Action} ([xshift=-0.4cm]env.north);

\draw[arrow] ([xshift=0.4cm]env.north) -- node[label style, right] {State, Reward} ([xshift=0.4cm]policy.south);

\end{tikzpicture}
    }
\caption{This diagram represents a sequential decision-making system in which an agent interacts with an environment. The agent receives a state and a reward from the environment based on its previous action. The agent then uses this information to select the next action, which it sends to the~environment.}
\label{fig:rl}
\end{figure}
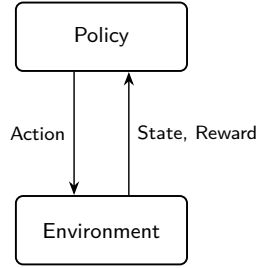

The interaction between the policy and environment is described in Figure~\ref{fig:rl}.

\subsection{Probabilistic Model Checking}

Storm~\cite{DBLP:journals/sttt/HenselJKQV22} is a model checker. 
It enables the verification of properties in induced DTMCs, with reachability properties being among the most fundamental.
These properties assess the probability of a system reaching a particular state.
For example, one might ask, ``Is the probability of the system reaching an unsafe state less than 0.1?''
A property can be either \emph{satisfied} or \emph{violated}.

\begin{figure}[htbp]
    \centering
    \scalebox{0.4}{
    \tikzset{
    ellipse node/.style={
        ellipse,
        draw=black,
        line width=0.8pt,
        fill=white,
        minimum width=2.2cm,
        minimum height=0.8cm,
        align=center,
        font=\sffamily\footnotesize
    },
    process block/.style={
        rectangle,
        rounded corners=3pt,
        draw=black,
        line width=0.8pt,
        fill=gray!15,
        minimum width=2.4cm,
        minimum height=0.9cm,
        align=center,
        font=\sffamily\small
    },
    arrow/.style={
        -{Stealth[length=2mm, width=1.5mm]},
        line width=0.6pt,
        color=black
    },
    label style/.style={
        font=\sffamily\scriptsize
    }
}

\begin{tikzpicture}[node distance=0.7cm]

\node[ellipse node] (system) {system};
\node[ellipse node, below=0.6cm of system] (systemdesc) {system description};
\node[ellipse node, below=0.6cm of systemdesc] (model) {model};

\node[ellipse node, right=1.2cm of systemdesc] (requirements) {requirements};
\node[ellipse node, below=0.6cm of requirements] (properties) {properties};

\node[process block, below=0.6cm of model, xshift=1.2cm] (modelchecking) {model checking};

\node[ellipse node, below left=0.6cm and 0.2cm of modelchecking] (satisfied) {satisfied};
\node[ellipse node, below right=0.6cm and 0.2cm of modelchecking] (violated) {violated};

\draw[arrow] (system) -- node[label style, right] {modeling} (systemdesc);
\draw[arrow] (systemdesc) -- node[label style, right] {translates to} (model);
\draw[arrow] (requirements) -- node[label style, right] {formalizing} (properties);
\draw[arrow] (model) -- (modelchecking);
\draw[arrow] (properties) -- (modelchecking);
\draw[arrow] (modelchecking) -- (satisfied);
\draw[arrow] (modelchecking) -- (violated);

\end{tikzpicture}
    }
    \caption{General model checking workflow~\cite{DBLP:journals/sttt/HenselJKQV22}. First, the system needs to be formally modeled, for instance, via PRISM. Then, the requirements are formalized, for instance, via PCTL. Eventually, both are inputted into the model checker, like Storm, which verifies the property.}
    \label{fig:model_checking}
\end{figure}
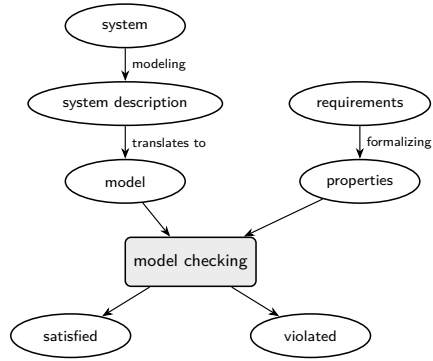

The \emph{general workflow} for model checking with Storm is as follows (see also Figure~\ref{fig:model_checking}):
First, the system, in our setting, an induced DTMC, is modeled using a language such as PRISM~\cite{prism_manual}.
Next, a property is formalized based on the system's requirements.
Using these inputs, the model checker Storm verifies whether the formalized property holds or fails within the model.

In probabilistic model checking, there is no universal ``one-size-fits-all'' solution~\cite{DBLP:journals/sttt/HenselJKQV22}.
The most suitable tools and techniques depend on the specific input model and properties being analyzed.
During model checking, Storm can proceed ``on the fly'', exploring only the parts of the DTMC most relevant to the formal verification.

Properties verifiable via Storm include temporal logic formulas for DTMCs, defined on paths using PCTL~\cite{hansson1994logic}, a branching-time logic.

Although it is not formally allowed in PCTL, Storm can still be used to request the probability of fulfilling a path formula from each state.
Rather than simply checking whether certain PCTL properties, such as $P_{=1}(\text{F } \textit{collision})$, are satisfied, which would indicate that the system reaches the $collision$ state with probability 1, we can query Storm to compute the actual probability value, denoted as $P(\text{F } \textit{collision})$, which in this case equals 1.

In this paper, we consider the \emph{eventually} path property $\text{F }$, which states that a property holds at some future state along a path. Its syntax is
\[
P_{\sim p} (\text{F } \varphi)
\]
meaning that the probability of reaching a state where $\varphi$ holds, along a path starting from the initial state, satisfies the bound $\sim p$, where $\sim$ is a comparison operator such as $<$, $\leq$, $\geq$, or $>$, and $p$ is a probability threshold.

Beyond verification, Storm can also synthesize optimal policies for MDPs.
Given an MDP and a property specification, Storm computes a policy that maximizes or minimizes the probability of satisfying the property.
This capability allows us to extract an optimal policy for a specified objective, which can then serve as the expert dataset for behavioral cloning, providing its state-action pairs~\cite{DBLP:conf/setta/Gross22}.

Assume the state space has some structure $S \subseteq Q \times I$ for suitable $Q$ and $I$. Thus,  states are pairs $s = (q,i)$ where $q$ and $i$ correspond to \emph{features} whose values range over $Q$ and $I$ respectively. Let $\textsf{Act}$ denote the actions in the MDP.
Given a partition $K_1,\dots,K_n$ of $I$, we abstract this feature from the policy $\pi$, by defining a \emph{permissive policy}~\cite{DBLP:journals/corr/DragerFK0U15} $\tau \colon S \rightarrow 2^\textsf{Act}$, i.e., a policy selecting multiple actions in every state. In particular, we consider $\tau(q,i) = \bigcup_{k \in K_n} \pi(q,k)$, with $K_n$ is the unique set such that $i \in K_n$.
This policy $\tau$ ignores the value of $i$ in state $(q,i)$ and instead selects any action that can be selected from states $(q,k)$, with $k \in K_n$.
Applying a permissive policy yields an induced MDP, which can be model checked to provide best- and worst-case~bounds.

\subsection{Behavioral Cloning}
\emph{Behavioral cloning} is an imitation learning approach that frames policy learning as supervised classification~\cite{DBLP:journals/corr/abs-2204-05618}. Given a dataset of expert demonstrations consisting of state-action pairs, behavioral cloning trains a policy to predict the expert's action for each observed state.
The policy is optimized using standard cross-entropy loss, treating action prediction as a classification problem. A key advantage of behavioral cloning is its simplicity: it reduces sequential decision-making to supervised learning, enabling the use of well-established optimization techniques.
However, behavioral cloning assumes access to high-quality expert demonstrations and can suffer from distribution shift, where errors compound as the learned policy visits states not represented in the training data.

\subsection{Saliency Maps for Feature Importance}
\emph{Saliency maps} are a gradient-based method for explaining neural network predictions by quantifying the importance of each input feature~\cite{DBLP:conf/aiccsa/DakhliB24}. The core idea is to measure how much a small change in each input feature would affect the network's output. Features that strongly influence the output receive higher importance scores. By aggregating these scores across all inputs, we obtain a global feature-importance ranking that characterizes which aspects of the inputs the network relies on most heavily. 

\section{Methodolodgy}
\begin{algorithm}[t]
\caption{COOL-MC: Formal Verification of Policies}
\label{alg:qverifier}
\begin{algorithmic}[1]
\Require MDP $M = (S, s_0, Act, Tr, rew, AP, L)$
\Require Trained policy $\pi$
\Require PCTL property $\varphi$
\Ensure Satisfaction result and probability $p$

\Statex
\Statex \textbf{Stage 1: Induced DTMC Construction}
\State $M^\pi \gets (S^\pi, s_0, Tr^\pi, AP, L)$ where $S^\pi \gets \emptyset$, $Tr^\pi \gets \emptyset$
\State \textsc{BuildDTMC}($s_0$)

\Statex
\Statex \textbf{Stage 2: Probabilistic Model Checking}
\State $(result, p) \gets \textsc{Storm}.\text{verify}(M^{\pi}, \varphi)$
\State \Return $(result, p)$

\Statex
\Procedure{BuildDTMC}{$s$}
    \If{$s \in S^\pi$ \textbf{or} $s$ is not relevant for $\varphi$}
        \State \Return
    \EndIf
    \State $S^\pi \gets S^\pi \cup \{s\}$
    \State $a \gets \pi(s)$
    \ForAll{$s' \in S$ where $Tr(s, a, s') > 0$}
        \State $Tr^\pi(s, s') \gets Tr(s, a, s')$
        \State \textsc{BuildDTMC}($s'$)
    \EndFor
\EndProcedure

\end{algorithmic}
\end{algorithm}
\emph{The Rashomon Effect in Sequential Decision-making} describes the observation that multiple policies trained on the same data select the same action in every state in an environment concerning the same property, thereby producing identical induced DTMCs, while differing in their internal structure as defined by a user-specified metric, such as feature attributions.
This is analogous to the classification Rashomon effect, where models make identical predictions while differing internally; here, identical induced DTMCs concerning a specified objective are synonymous with identical predictions (see visual comparison in Figure~\ref{fig:classifiers}).

The set of policies exhibiting this effect forms the \emph{Rashomon set}.
Policies that are identical in both observable behavior and internal structure under the specified metric do not constitute a Rashomon set.

In classification and regression, verifying identical behavior is straightforward: one compares predictions to ground truth.
In sequential decision-making with stochastic transitions, this simple comparison to the ground truth is not straightforward, as the same policy may succeed or fail on any single trajectory due to the stochastic nature of the environment dynamics.
Probabilistic model checking resolves this by explicitly constructing and comparing the induced DTMCs, whereas user-specified explainable machine learning methods verify that the internal structures differ.

In the following subsections, we outline the methodology to construct the Rashomon set.
First, all policies must be trained on an identical expert dataset.
Second, probabilistic model checking identifies which policies produce identical induced DTMCs for a specified property; these are candidates for the Rashomon set.
Third, user-specified explainable machine learning methods, such as saliency-based feature rankings, verify that candidate policies differ in their internal structure.
Finally, only policies that satisfy all criteria are included in the Rashomon set; policies that are identical in both behavior and internal structure are excluded.

\subsection{Policy Training}
To ensure that all policies are trained on identical data, we require a fixed expert dataset.
In our setting, we obtain this dataset by first extracting an optimal policy $\pi^*$ concerning an objective/property using the Storm model checker; alternatively, any method that produces fixed training data, such as offline reinforcement learning~\cite{DBLP:journals/corr/abs-2204-05618}, could be used.

Given a formal environment model specified as an MDP and a property specification, Storm synthesizes a policy that maximizes the probability of satisfying the specified property.
From this optimal policy, we collect all state-action pairs for all states in the MDP to construct the expert dataset $\mathcal{D} = \{(s_i, a_i)\}_{i=1}^{N}$, where each pair represents the optimal action $a_i$ to take in state $s_i$.

We train multiple neural network policies on the expert dataset using behavioral cloning.
Each policy is initialized with a different random seed, leading to different weight initializations and optimization trajectories.
This mirrors the classification setting in which multiple models are trained on identical labeled data, thereby maintaining the core criterion of the Rashomon~effect.

Note: training accuracy is not our metric of interest; rather, we evaluate policies based on their induced behavior as described in the next subsection.

\subsection{Verifying Identical Induced DTMCs}
To confirm that policies exhibit identical observable behavior, we use COOL-MC~\cite{DBLP:conf/setta/Gross22} to construct and compare their induced DTMCs.
For each trained policy $\pi$, property, and the formal environment model $M$, COOL-MC incrementally constructs the induced Markov chain that captures the complete interaction between the policy and the environment concerning the specified property.

Starting from the initial state, COOL-MC queries the policy for an action $a = \pi(s)$ at each reachable state $s$ and expands only those successor states $s'$ that are reachable via that action in the underlying MDP (see Algorithm~\ref{alg:qverifier}).

\subsection{Revealing Internal Policy Differences}
Once we have identified candidate policies with identical induced DTMCs, we investigate whether they differ internally using a user-specified explainable machine learning method.
While any suitable method can be used to characterize internal structure, in this paper we focus on saliency-based feature attribution.

For each policy $\pi_i$, we compute a feature importance ranking $\rho_i$ by aggregating gradient-based attribution scores across all states in the dataset.
The Rashomon effect manifests when policies with identical induced DTMCs exhibit different feature importance rankings.
For example, policy $\pi_1$ might rank fuel as most important while policy $\pi_2$ ranks spatial coordinates as most important, even though both policies select identical actions in every state (see Definition~\ref{def:rashomon_effect}).

\begin{definition}[Rashomon Effect in Sequential Decision-Making]\label{def:rashomon_effect}
Let $\Pi = \{\pi_1, \ldots, \pi_n\}$ be a set of policies trained on the same expert dataset $\mathcal{D}$, let $M$ be an MDP, let $\varphi$ be a property specification, and let $\delta$ be a user-specified metric for internal structure (e.g., feature attribution rankings). The \emph{Rashomon effect in sequential decision-making} occurs when there exist policies $\pi_i, \pi_j \in \Pi$ such that:
\begin{enumerate}
    \item $\pi_i$ and $\pi_j$ induce identical DTMCs on $M$ with respect to $\varphi$, and
    \item $\delta(\pi_i) \neq \delta(\pi_j)$.
\end{enumerate}
\end{definition}

\subsection{Creating the Rashomon Set}
From the candidate policies with identical induced DTMCs, we include in the Rashomon set only those policies that differ in their internal structure under the specified metric.
Policies that are identical in both observable behavior and internal structure under the specified metric are excluded from the Rashomon set (see Definition~\ref{def:rashomon_set}).

\begin{definition}[Rashomon Set]\label{def:rashomon_set}
Given a set of policies $\Pi$ trained on the same expert dataset $\mathcal{D}$, an MDP $M$, a property specification $\varphi$, and an internal structure metric $\delta$, a \emph{Rashomon set} $R \subseteq \Pi$ is a subset such that every pair of distinct policies $\pi_i, \pi_j \in R$ exhibits the Rashomon effect (Definition~\ref{def:rashomon_effect}).
\end{definition}

\section{Experiments}
We conduct experiments to validate the existence of the Rashomon effect in sequential decision-making and to evaluate its practical implications. Specifically, we investigate the following two research questions:
\begin{itemize}
    \item \textbf{RQ1:} \emph{Does the Rashomon effect exist in sequential decision-making?} Specifically, can multiple policies trained on the same dataset produce identical induced DTMCs while exhibiting different feature importance rankings?
    \item \textbf{RQ2:} \emph{Do internal differences manifest under distribution shift?} Specifically, do policies from the Rashomon set, which behave identically in the original environment, diverge in behavior when the environment changes?
\end{itemize}
Before we answer these research questions, we describe the environment, policy training procedure, and technical setup.

\subsection{Setup}
In this subsection, we describe the setup of our experiments that will answer our research questions.
\subsubsection*{Environment}
We evaluate our approach using a \emph{taxi} environment.
The taxi agent must pick up passengers and transport them to their destination without running out of fuel.
The environment terminates when the taxi completes a predefined number of jobs or runs out of fuel.
After each completed job, a new passenger spawns randomly at one of four predefined locations~\cite{DBLP:conf/setta/Gross22}.
For the first job, the passenger location and destination are fixed; for subsequent jobs, both are assigned randomly.
The state space, action space, and reward are defined as~follows:
\begin{gather*}
S = \{(x, y, passenger\_loc\_x, passenger\_loc\_y, passenger\_dest\_x, passenger\_dest\_y,\\ fuel, on\_board, jobs\_done, done), ...\} \\
Act = \{north, east, south, west, pick\_up, drop\}\\
reward = \text{Not needed due to behavioral cloning.}
\end{gather*}

\subsubsection*{Policy training}
To ensure that all policies are trained on the same dataset, we use behavioral cloning.
We first use the Storm model checker to extract the optimal policy for the property of completing five jobs in the taxi environment.
From this optimal policy, we collect the state-action pairs for all states in the full MDP to construct the expert dataset.
We then train multiple neural network policies on this dataset with different random seeds.
This setup mirrors the supervised learning setting where multiple models are trained on identical data, enabling a direct investigation of the Rashomon effect.

\subsubsection*{Technical setup}
We executed our benchmarks in a Docker container with 16 GB RAM, and an AMD Ryzen 7 7735hs with Radeon graphics × 16 processor with the operating system Ubuntu 20.04.5 LTS.
For model checking, we use Storm 1.7.0.

\subsection{Analysis}
In this subsection, we answer the previously mentioned research questions.


\definecolor{groupA}{HTML}{FFD8D8}
\definecolor{groupB}{HTML}{D8D8FF}
\definecolor{groupC}{HTML}{D8EBD8}
\definecolor{groupD}{HTML}{FFFFCC}
\definecolor{groupE}{HTML}{FFF1D8}
\definecolor{groupF}{HTML}{EBD8EB}
\definecolor{groupG}{HTML}{D8FFFF}
\definecolor{groupH}{HTML}{EDE3DB}
\definecolor{groupI}{HTML}{FFF2F4}
\definecolor{groupJ}{HTML}{D8FFD8}

\begin{table}[h]
\centering
\caption{Feature importance rankings for the policies. Rankings range from 1 (most important) to 10 (least important). The MC column shows the model checking result for each policy. Rows with identical colors indicate policies with identical feature rankings. Despite identical induced DTMCs, the policies may emphasize different features.}
\label{tab:feature_rankings}
\begin{tabular}{lccccccccccc}
\toprule
Policy & $done$ & $fuel$ & $jd$ & $pas$ & $pd_x$ & $pd_y$ & $pl_x$ & $pl_y$ & $x$ & $y$ & MC \\
\midrule
\rowcolor{groupC}
$\pi_{15}$ & 1 & 2 & 7 & 5 & 9 & 6 & 10 & 8 & 3 & 4 & 1.0 \\
\rowcolor{groupC}
$\pi_{5}$ & 1 & 2 & 7 & 5 & 9 & 6 & 10 & 8 & 3 & 4 & 1.0 \\
\rowcolor{groupC}
$\pi_{50}$ & 1 & 2 & 7 & 5 & 9 & 6 & 10 & 8 & 3 & 4 & 1.0 \\
\rowcolor{groupH}
$\pi_{17}$ & 1 & 3 & 7 & 5 & 10 & 6 & 9 & 8 & 2 & 4 & 1.0 \\
\rowcolor{groupH}
$\pi_{63}$ & 1 & 3 & 7 & 5 & 10 & 6 & 9 & 8 & 2 & 4 & 1.0 \\
\rowcolor{groupH}
$\pi_{75}$ & 1 & 3 & 7 & 5 & 10 & 6 & 9 & 8 & 2 & 4 & 1.0 \\
\rowcolor{groupI}
$\pi_{19}$ & 1 & 3 & 7 & 5 & 9 & 8 & 10 & 6 & 2 & 4 & 1.0 \\
\rowcolor{groupI}
$\pi_{35}$ & 1 & 3 & 7 & 5 & 9 & 8 & 10 & 6 & 2 & 4 & 1.0 \\
\rowcolor{groupI}
$\pi_{77}$ & 1 & 3 & 7 & 5 & 9 & 8 & 10 & 6 & 2 & 4 & 1.0 \\
\rowcolor{groupA}
$\pi_{1}$ & 1 & 3 & 6 & 5 & 10 & 9 & 8 & 7 & 2 & 4 & 1.0 \\
\rowcolor{groupA}
$\pi_{51}$ & 1 & 3 & 6 & 5 & 10 & 9 & 8 & 7 & 2 & 4 & 1.0 \\
\rowcolor{groupE}
$\pi_{11}$ & 1 & 3 & 6 & 5 & 9 & 8 & 10 & 7 & 2 & 4 & 1.0 \\
\rowcolor{groupE}
$\pi_{86}$ & 1 & 3 & 6 & 5 & 9 & 8 & 10 & 7 & 2 & 4 & 1.0 \\
\rowcolor{groupG}
$\pi_{14}$ & 1 & 3 & 7 & 5 & 10 & 9 & 8 & 6 & 2 & 4 & 1.0 \\
\rowcolor{groupG}
$\pi_{47}$ & 1 & 3 & 7 & 5 & 10 & 9 & 8 & 6 & 2 & 4 & 1.0 \\
\rowcolor{groupJ}
$\pi_{20}$ & 1 & 3 & 7 & 5 & 10 & 8 & 9 & 6 & 2 & 4 & 1.0 \\
\rowcolor{groupJ}
$\pi_{49}$ & 1 & 3 & 7 & 5 & 10 & 8 & 9 & 6 & 2 & 4 & 1.0 \\
\rowcolor{groupB}
$\pi_{4}$ & 1 & 2 & 7 & 5 & 9 & 6 & 10 & 8 & 4 & 3 & 1.0 \\
\rowcolor{groupB}
$\pi_{54}$ & 1 & 2 & 7 & 5 & 9 & 6 & 10 & 8 & 4 & 3 & 1.0 \\
\rowcolor{groupD}
$\pi_{41}$ & 1 & 3 & 8 & 5 & 9 & 7 & 10 & 6 & 4 & 2 & 1.0 \\
\rowcolor{groupD}
$\pi_{7}$ & 1 & 3 & 8 & 5 & 9 & 7 & 10 & 6 & 4 & 2 & 1.0 \\

\bottomrule
\end{tabular}%

\vspace{0.5em}

\small{$jd$ = jobs\_done, $pas$ = passenger, $pd_x$ = passenger\_dest\_x, $pd_y$ = passenger\_dest\_y, $pl_x$ = passenger\_loc\_x, $pl_y$ = passenger\_loc\_y}

\end{table}

\subsubsection*{RQ1: Does the Rashomon effect exist in sequential decision-making?}
We trained 100 policies via behavioral cloning on the expert dataset, each initialized with a different random seed.

Using COOL-MC, we constructed the induced DTMC for each policy concerning the property of finishing five jobs with a reachability probability of 1.
Comparing these induced DTMCs, we identified 10 behavioral equivalence classes, where policies in the same class exhibit identical behavior.
The largest class contains 82 policies that satisfy the property.
The remaining classes contain policies that fail to satisfy the property, ranging from never completing five jobs to achieving reachability probabilities between zero and one.

We focus our analysis on the 82 policies in the largest behavioral equivalence class.
Within this class, we computed feature attributions for each policy and identified attribution patterns, where policies with the same pattern assign identical importance rankings to all features.

Table~\ref{tab:feature_rankings} shows a subset of these policies. Policies sharing the same row color have identical attribution patterns and therefore cannot form a Rashomon set together.
For space reasons, we show only policies that share their attribution pattern with at least one other policy.

This experiment demonstrates that the Rashomon effect exists in sequential decision-making: policies can be behaviorally identical yet differ internally.

\begin{table}[h]
\centering
\caption{Policy behavior under distribution shift (increasing job count). Policies that produced identical induced DTMCs for 5 jobs diverge as the task becomes more demanding. Values show the probability of completing all jobs.}
\label{tab:rq2_results}
\begin{tabular}{lcccccc}
\toprule
$\pi$ & 5 jobs & 6 jobs & 7 jobs & 8 jobs & 9 jobs & 10 jobs \\
\midrule
$\pi_1$ & 1.0 & 1.0 & 0.0 & 0.0 & 0.0 & 0.0 \\
$\pi_2$ & 1.0 & 1.0 & 1.0 & 0.23 & 0.0 & 0.0 \\
$\pi_3$ & 1.0 & 0.5 & 0.0 & 0.0 & 0.0 & 0.0 \\
$\pi_4$ & 1.0 & 0.875 & 0.176 & 0.0 & 0.0 & 0.0 \\
$\pi_5$ & 1.0 & 1.0 & 0.94 & 0.0 & 0.0 & 0.0 \\
\midrule
$\mathbb{E}[\pi_{1:5}]$ & 1.0 & 0.875 & 0.423 & 0.046 & 0.0 & 0.0 \\
\midrule
$\pi_R$ & 1.0 & 1.0 & 1.0 & 0.0 & 0.0 & 0.0 \\
$\tau_R$ & 1.0 & 1.0 & 1.0 & 1.0 & 1.0 & 1.0 \\
\midrule
$\pi^*$ & 1.0 & 1.0 & 1.0 & 1.0 & 1.0 & 1.0 \\
\bottomrule
\end{tabular}
\end{table}

\vspace{0.5em}
\begin{mdframed}
\textbf{Answer to RQ1:} Yes, the Rashomon effect exists in sequential decision-making.
\end{mdframed}

\subsubsection*{RQ2: Do internal differences manifest under distribution shift?}
Given a Rashomon set of five policies that behave identically in the original training environment, a natural question is whether their internal differences have any practical consequences.
We evaluated the Rashomon set policies under distribution shifts induced by increasing the number of jobs the agent must complete (6, 7, 8, 9, 10 jobs instead of 5).

Table~\ref{tab:rq2_results} shows the results.
Under distribution shift, the policies that were observationally identical in the original environment now produce different induced DTMCs and achieve different success probabilities.
The Rashomon ensemble $\pi_R$, constructed via majority voting, outperforms randomly selecting any individual policy, which would be the only option since all policies are indistinguishable in the original environment.
Therefore, the Rashomon ensemble saves time.

More notably, by constructing a permissive policy $\tau_R$ from the Rashomon set, which allows any action selected by at least one member policy, we obtain an induced MDP rather than a DTMC.
Model checking this induced MDP reveals that it still contains an optimal policy achieving 100\% success rate across all distribution shifts.
This approach, similar to~\cite{DBLP:conf/setta/Gross22}, still avoids constructing the full state space of the original MDP.
The induced MDP contains only 6,209 states and 12,149 transitions, compared to 72,064 states and 335,544 transitions in the original MDP, offering significant memory savings for model checking tasks where the full model is too large to load into memory.

\vspace{0.5em}
\begin{mdframed}
\textbf{Answer to RQ2:} Yes, internal differences manifest under distribution shift. Policies that were observationally identical for five jobs diverged significantly as the job count increased. 
The permissive Rashomon ensemble $\tau_R$ achieved 100\% success across all distribution shifts while reducing state space, outperforming both individual policies and the majority-voting Rashomon ensemble.
\end{mdframed}

\section{Discussion}
A central methodological contribution of this work is the use of probabilistic model checking to verify identical observable behavior among policies.
In classification and regression, one compares predictions against ground truth, but in sequential decision-making with stochastic transitions, the same policy may succeed or fail on any trajectory due to randomness.
A faithful translation of the Rashomon effect requires definitive equivalence (see Figure~\ref{fig:classifiers}), yet empirical success rates provide only statistical estimates.
Probabilistic model checking resolves this by exactly constructing the induced discrete-time Markov chain for each policy, enabling precise comparison and allowing us to isolate internal differences as the sole source of variation between~policies.

Beyond verification, the Rashomon set offers practical benefits for model checking itself. State space explosion remains a fundamental challenge, as full environment models can be too large to load into memory~\cite{DBLP:conf/setta/Gross22}.
By constructing a permissive policy from the Rashomon set, we obtain an induced MDP that is substantially smaller than the original while still containing an optimal policy, as demonstrated by the reduction from 72,064 states to 6,209 states in our experiments. Additionally, the permissive Rashomon ensemble outperforms randomly selecting any individual policy under distribution shift, providing robustness benefits when policies are otherwise indistinguishable in the original environment.
This approach also saves time compared to sequentially model checking individual policies with identical performance to find one that generalizes well.

Additionally, while we used behavioral cloning to ensure that all policies are trained on identical data, alternative approaches such as offline reinforcement learning~\cite{DBLP:journals/apin/ChenZY25} could also provide a common dataset.
In online reinforcement learning, where each policy accumulates its own experience, one could produce multiple policies by pruning the original trained neural network policy while preserving the original policy's behaviour~\cite{DBLP:conf/esann/GrossS24}.

Our methodology is agnostic to the specific metric used to characterize internal policy structure.
While we employed saliency-based feature attributions to reveal internal differences, other explainable machine learning methods could serve this purpose.
The choice of metric may influence which policies are included in the Rashomon set, as two policies might appear identical under one metric but differ under another.

The existence of the Rashomon effect in sequential decision-making has implications for explainable reinforcement learning.
When multiple policies achieve the same objective with identical observable behavior, explanations become underdetermined: different yet equally valid explanations exist for the same outcomes.
This raises questions about which explanation to~trust.

Finally, the Rashomon effect, as defined in this paper, applies to any domain that can be cast as sequential decision-making.
Classification and regression tasks can be viewed as one-step sequential decision-making tasks, whereas text generation involves multi-step action selection, with each token constituting an action.
More generally, any sequence-to-sequence task, such as machine translation, speech recognition, audio synthesis, video generation, or code generation,  involves sequential action selection that could exhibit the Rashomon effect.
The key requirements are: (1) identical training data, (2) a formal specification of identical observable behavior, and (3) a meaningful metric for internal structural differences.

\section{Conclusion}
In this paper, we defined and demonstrated the Rashomon effect for sequential decision-making.
Just as classification models in the Rashomon set make identical predictions while differing in feature importance, policies in our sequential decision-making Rashomon set produce identical induced DTMCs while differing in feature attributions (see Experiments and Figure~\ref{fig:classifiers} for a visual high-level example).

We verified the existence of this effect using probabilistic model checking to confirm identical induced DTMCs and saliency-based methods to reveal different feature attributions.
Furthermore, we showed that these internal differences manifest as divergent behavior under a distribution shift.
This provides an objective criterion to evaluate explanation quality when policies are otherwise indistinguishable.

\emph{Future work} could investigate the Rashomon effect in multi-agent systems~\cite{DBLP:conf/aips/GrossS0023}, explore how differences in feature attribution relate to specific types of distribution shifts, analyze the Rashomon ensemble in the context of probabilistic model checking~\cite{DBLP:conf/setta/Gross22}, and investigate how different internal structure metrics affect Rashomon set composition.

\bibliographystyle{plain}
\bibliography{refs}

@inproceedings{muller_empirical_2023,
  title={An empirical evaluation of the Rashomon effect in explainable machine learning},
  author={M{\"u}ller, Sebastian and Toborek, Vanessa and Beckh, Katharina and Jakobs, Matthias and Bauckhage, Christian and Welke, Pascal},
  booktitle={Joint European Conference on Machine Learning and Knowledge Discovery in Databases},
  pages={462--478},
  year={2023},
  organization={Springer}
}

@inproceedings{Xin2022,
  title = {Exploring the Whole Rashomon Set of Sparse Decision Trees},
  author = {Rui Xin and Chudi Zhong and Zhi Chen and Takuya Takagi and Margo I. Seltzer and Cynthia Rudin},
  year = {2022},
  month = {dec},
  journal = {Advances in Neural Information Processing Systems},
  volume = {35},
  pages = {14071--14084},
  timestamp = {Mon, 08 Jan 2024 16:31:37 +0100},
  biburl = {https://dblp.org/rec/conf/nips/XinZ0TSR22.bib},
  bibsource = {dblp computer science bibliography, https://dblp.org},
  url = {http://papers.nips.cc/paper\_files/paper/2022/hash/5afaa8b4dd18eb1eed055d2d821b58ae-Abstract-Conference.html},
  booktitle = {Advances in Neural Information Processing Systems 35: Annual Conference on Neural Information Processing Systems 2022, NeurIPS 2022, New Orleans, LA, USA, November 28 - December 9, 2022},
  _bib2doi_selected = {dblp:/rec/conf/nips/XinZ0TSR22.bib},
  _bib2doi_confirmed = {true},
  _bib2doi_finished = {true},
}

@article{hansson1994logic,
  title={A logic for reasoning about time and reliability},
  author={Hansson, Hans and Jonsson, Bengt},
  journal={Formal aspects of computing},
  volume={6},
  number={5},
  pages={512--535},
  year={1994},
  publisher={Springer}
}

@article{DBLP:journals/corr/abs-2509-03169,
  author       = {Helge Spieker and
                  J{\o}rn Eirik Betten and
                  Arnaud Gotlieb and
                  Nadjib Lazaar and
                  Nassim Belmecheri},
  title        = {Rashomon in the Streets: Explanation Ambiguity in Scene Understanding},
  journal      = {CoRR},
  volume       = {abs/2509.03169},
  year         = {2025}
}

@inproceedings{DBLP:conf/aiccsa/DakhliB24,
  author       = {Rym Dakhli and
                  Walid Barhoumi},
  title        = {Toward a Quantitative Trustworthy Evaluation of Post-Hoc {XAI} Feature
                  Importance Maps Using Saliency-Based Occlusion},
  booktitle    = {{AICCSA}},
  pages        = {1--8},
  publisher    = {{IEEE}},
  year         = {2024}
}

@inproceedings{DBLP:conf/setta/Gross22,
  author    = { Dennis Gross and
                Nils Jansen and
                Sebastian Junges and
                Guillermo A. P{\'e}rez
                },
  title     = {{COOL-MC}: A Comprehensive Tool for
Reinforcement Learning and Model Checking},
  booktitle = {{SETTA}},
  publisher = {Springer},
  year      = {2022}
}

@article{DBLP:journals/corr/DragerFK0U15,
  author    = {Klaus Dr{\"{a}}ger and
               Vojtech Forejt and
               Marta Z. Kwiatkowska and
               David Parker and
               Mateusz Ujma},
  title     = {Permissive Controller Synthesis for Probabilistic Systems},
  journal   = {Log. Methods Comput. Sci.},
  volume    = {11},
  number    = {2},
  year      = {2015}
}

@inproceedings{delgrange2023wae,
  title={WAE-PCN: Wasserstein-autoencoded Pareto Conditioned Networks},
  author={Delgrange, Florent and Reymond, Mathieu and Now{\'e}, Ann and P{\'e}rez, Guillermo A},
  booktitle={2023 adaptive and learning agents workshop at AAMAS},
  pages={1--7},
  year={2023}
}

@article{DBLP:journals/iandc/LarsenS91,
  author       = {Kim Guldstrand Larsen and
                  Arne Skou},
  title        = {Bisimulation through Probabilistic Testing},
  journal      = {Inf. Comput.},
  volume       = {94},
  number       = {1},
  pages        = {1--28},
  year         = {1991}
}

@article{ferns2011bisimulation,
  title={Bisimulation metrics for continuous Markov decision processes},
  author={Ferns, Norm and Panangaden, Prakash and Precup, Doina},
  journal={SIAM Journal on Computing},
  volume={40},
  number={6},
  pages={1662--1714},
  year={2011},
  publisher={SIAM}
}

@article{DBLP:journals/apin/ChenZY25,
  author       = {Jiacheng Chen and
                  Jin Zhu and
                  Lin Yang},
  title        = {Offline-to-online reinforcement learning with policy ensemble and
                  policy-extended value},
  journal      = {Appl. Intell.},
  volume       = {55},
  number       = {16},
  pages        = {1067},
  year         = {2025}
}

@misc{prism_manual, 
title="{PRISM Manual}",
url={https://www.prismmodelchecker.org/doc/semantics.pdf},
author = "{PRISM}",
year = "{2023}",
howpublished = "\url{www.prismmodelchecker.org}",
note = "Accessed: {03/14/2024}"
}

@inproceedings{DBLP:conf/aips/GrossS0023,
  author       = {Dennis Gross and
                  Christoph Schmidl and
                  Nils Jansen and
                  Guillermo A. P{\'{e}}rez},
  title        = {Model Checking for Adversarial Multi-Agent Reinforcement Learning
                  with Reactive Defense Methods},
  booktitle    = {{ICAPS}},
  pages        = {162--170},
  publisher    = {{AAAI} Press},
  year         = {2023}
}

@article{DBLP:journals/jair/NashedMGZ25,
  author       = {Samer B. Nashed and
                  Saaduddin Mahmud and
                  Claudia V. Goldman and
                  Shlomo Zilberstein},
  title        = {Causal Explanations for Sequential Decision Making},
  journal      = {J. Artif. Intell. Res.},
  volume       = {83},
  year         = {2025}
}

@inproceedings{DBLP:conf/atva/Gross0PR20,
  author       = {Dennis Gross and
                  Nils Jansen and
                  Guillermo A. P{\'{e}}rez and
                  Stephan Raaijmakers},
  title        = {Robustness Verification for Classifier Ensembles},
  booktitle    = {{ATVA}},
  series       = {Lecture Notes in Computer Science},
  volume       = {12302},
  pages        = {271--287},
  publisher    = {Springer},
  year         = {2020}
}

@inproceedings{DBLP:conf/esann/GrossS24,
  author       = {Dennis Gross and
                  Helge Spieker},
  title        = {Safety-Oriented Pruning and Interpretation of Reinforcement Learning
                  Policies},
  booktitle    = {{ESANN}},
  year         = {2024}
}

@article{DBLP:journals/corr/abs-2204-05618,
  author       = {Aviral Kumar and
                  Joey Hong and
                  Anikait Singh and
                  Sergey Levine},
  title        = {When Should We Prefer Offline Reinforcement Learning Over Behavioral
                  Cloning?},
  journal      = {CoRR},
  volume       = {abs/2204.05618},
  year         = {2022}
}

@inproceedings{DBLP:conf/icoin/ShinK21,
  author       = {Myungjae Shin and
                  Joongheon Kim},
  title        = {Joint Behavioral Cloning and Reinforcement Learning Method for Propofol
                  and Remifentanil Infusion in Anesthesia},
  booktitle    = {{ICOIN}},
  pages        = {849--852},
  publisher    = {{IEEE}},
  year         = {2021}
}

@inproceedings{DBLP:conf/icaart/SchmidlS024,
  author       = {Christoph Schmidl and
                  Thiago D. Sim{\~{a}}o and
                  Nils Jansen},
  title        = {A Supervised Learning Approach to Robust Reinforcement Learning for
                  Job Shop Scheduling},
  booktitle    = {{ICAART} {(3)}},
  pages        = {1324--1335},
  publisher    = {{SCITEPRESS}},
  year         = {2024}
}

@book{sutton2018reinforcement,
  title={Reinforcement learning: An introduction},
  author={Sutton, Richard S and Barto, Andrew G},
  year={2018},
  publisher={MIT press}
}

@article{Laberge2023,
  title = {Partial {{Order}} in {{Chaos}}: {{Consensus}} on {{Feature Attributions}} in the {{Rashomon Set}}},
  shorttitle = {Partial {{Order}} in {{Chaos}}},
  author = {Laberge, Gabriel and Pequignot, Yann and Mathieu, Alexandre and Khomh, Foutse and Marchand, Mario},
  year = {2023},
  journal = {Journal of Machine Learning Research},
  volume = {24},
  number = {364},
  _bib2doi_finished = {true},
}

@inproceedings{Andersen2023,
  ids = {Andersen2023},
  title = {Producing {{Diverse Rashomon Sets}} of {{Counterfactual Explanations}} with {{Niching Particle Swarm Optimization Algorithms}}},
  booktitle = {Proceedings of the {{Genetic}} and {{Evolutionary Computation Conference}}},
  author = {Andersen, Hayden and Lensen, Andrew and Browne, Will and Mei, Yi},
  year = {2023},
  series = {{{GECCO}} '23},
  pages = {393--401},
  publisher = {Association for Computing Machinery},
  address = {New York, NY, USA},
  doi = {10.1145/3583131.3590444},
  isbn = {9798400701191},
  timestamp = {Mon, 01 Apr 2024 11:14:08 +0200},
  biburl = {https://dblp.org/rec/conf/gecco/AndersenLB023.bib},
  bibsource = {dblp computer science bibliography, https://dblp.org},
  _bib2doi_selected = {dblp:/rec/conf/gecco/AndersenLB023.bib},
  _bib2doi_confirmed = {true},
}

@inproceedings{semenova_existence_2022,
	title = {On the {Existence} of {Simpler} {Machine} {Learning} {Models}},
	url = {http://arxiv.org/abs/1908.01755},
	doi = {10.1145/3531146.3533232},
	urldate = {2024-03-19},
	booktitle = {2022 {ACM} {Conference} on {Fairness}, {Accountability}, and {Transparency}},
	author = {Semenova, Lesia and Rudin, Cynthia and Parr, Ronald},
	month = jun,
	year = {2022},
	note = {arXiv:1908.01755 [cs, stat]},
	pages = {1827--1858},
}

@book{baier2008principles,
  title     = {Principles of model checking},
  author    = {Baier, Christel and Katoen, Joost-Pieter},
  year      = {2008},
  publisher = {MIT press}
}

@article{DBLP:journals/sttt/HenselJKQV22,
  author    = {Christian Hensel and
               Sebastian Junges and
               Joost{-}Pieter Katoen and
               Tim Quatmann and
               Matthias Volk},
  title     = {The probabilistic model checker {Storm}},
  journal   = {Int. J. Softw. Tools Technol. Transf.},
  volume    = {24},
  number    = {4},
  pages     = {589--610},
  year      = {2022}
}

@book{goodfellow2016deep,
title={Deep learning},
author={Goodfellow, Ian and Bengio, Yoshua and Courville, Aaron and Bengio, Yoshua},
volume={1},
year={2016},
publisher={MIT Press}
}

@article{breiman2001statistical,
  title={Statistical modeling: The two cultures (with comments and a rejoinder by the author)},
  author={Breiman, Leo},
  journal={Statistical science},
  volume={16},
  number={3},
  pages={199--231},
  year={2001},
  publisher={Institute of Mathematical Statistics}
}

@article{fisher_all_2019,
	title = {All {Models} are {Wrong}, but {Many} are {Useful}: {Learning} a {Variable}’s {Importance} by {Studying} an {Entire} {Class} of {Prediction} {Models} {Simultaneously}},
	volume = {20},
	issn = {1532-4435},
	shorttitle = {All {Models} are {Wrong}, but {Many} are {Useful}},
	url = {https://www.ncbi.nlm.nih.gov/pmc/articles/PMC8323609/},
	urldate = {2024-03-19},
	journal = {Journal of machine learning research : JMLR},
	author = {Fisher, Aaron and Rudin, Cynthia and Dominici, Francesca},
	year = {2019},
	pmid = {34335110},
	pmcid = {PMC8323609},
	pages = {177},
}

@inproceedings{DBLP:conf/icitjo/AbdiMWKASASA25,
  author       = {Abdinasir Hirsi Abdi and
                  Jafar Ismail Mohamed and
                  Abubakar Abdi Warsame and
                  Abdirizak Abdullahi Khalif and
                  Lukman Audah and
                  Adeb Salh and
                  Salman Ahmed and
                  Diani Galih Saputri and
                  Abdiaziz Mohamed Abdirahman},
  title        = {A Survey of Supervised, Unsupervised, and Ensemble Learning Approaches
                  for DDoS Detection in {SDN}},
  booktitle    = {{ICIT}},
  pages        = {99--104},
  publisher    = {{IEEE}},
  year         = {2025}
}

@inproceedings{DBLP:conf/glvlsi/Hu22,
  author       = {Jingtong Hu},
  title        = {Session details: Session 3B: {VLSI} for Machine Learning and Artifical
                  Intelligence 1},
  booktitle    = {{ACM} Great Lakes Symposium on {VLSI}},
  publisher    = {{ACM}},
  year         = {2022}
}

\end{document}